\documentclass{article}

\usepackage{graphicx}
\usepackage{wrapfig}
\usepackage{needspace}
\usepackage{tabularx}
\usepackage{setspace}
\usepackage{booktabs}
\usepackage{array}
\usepackage{amssymb}
\usepackage{mathtools}

\usepackage[final]{corl_2026} %

\title{YOCO: You Only Calibrate Once! \\ Fast Mocap Calibration for Dexterous Teleoperation}

\author{
  Yu Zhang$^{1,2,3}$, Yunqi Li$^{1,3}$, Yushi Du$^1$, Yi Ma$^{1,2,3}$, Yanchao Yang$^{1,2,3}$\\
  $^1$The University of Hong Kong, $^2$Shenzhen Loop Area Institute, $^3$TranscEngram
}

\newcommand{\MethodName}{\textit{YOCO}}

\begin{document}
\maketitle

\begin{figure}[ht]
    \centering
    \includegraphics[width=0.99\linewidth]{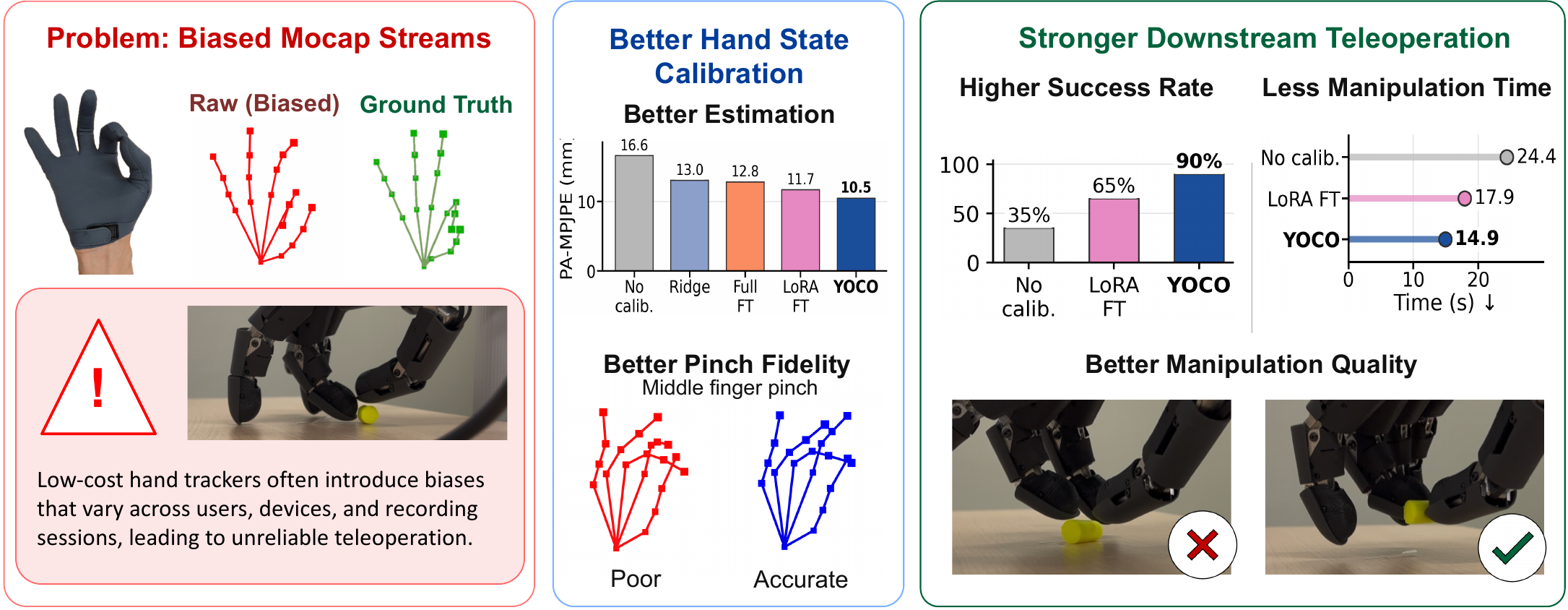}
    \caption{
    Overview of \MethodName{}. Low-cost mocap streams suffer from user-, device-, and session-specific bias. Given a few paired calibration poses, \MethodName{} predicts a calibration update that corrects future mocap poses and improves downstream dexterous teleoperation.
    }
    \label{fig:teaser}
\end{figure}

\begin{abstract}
Dexterous teleoperation requires reliable human-hand state estimations. However, common low-cost motion-capture gloves and markerless trackers often exhibit biases that vary across users, glove fit, and recording sessions, degrading retargeting and demonstration quality. We present \MethodName{}, a fast few-shot, fine-tuning-free calibration framework that corrects biased hand-pose streams from a small set of paired raw and target poses. Instead of optimizing a separate model for every operator or session, \MethodName{} conditions a calibration HyperNet on the paired examples and predicts LoRA-style updates for a frozen MANO hand-estimation module, turning per-user calibration into a lightweight feed-forward adaptation step while preserving the geometric prior of MANO and the efficiency of a compact estimator. We train \MethodName{} with synthetic drift augmentations on InterHand2.6M and evaluate on augmented InterHand sequences, offline real glove data, and dexterous teleoperation tasks. Across these settings, \MethodName{} improves calibration efficiency, hand-state estimation quality and teleoperation performance compared with uncalibrated input and standard calibration baselines.
\end{abstract}

\section{Introduction}
Teleoperation is a key tool for collecting dexterous hand demonstrations, but its effectiveness depends on the fidelity of the operator's hand-state estimates. High-end motion-capture systems can provide accurate kinematics, yet they are expensive and often difficult to deploy at scale. Lower-cost gloves and markerless trackers are more accessible, but their outputs are affected by user hand morphology gap, device fit, sensor drift, and session-dependent bias. Such errors are particularly detrimental in dexterous manipulation, where even small fingertip or joint offsets can alter contact patterns, pinch geometry, and retargeted robot motion.

Prior hand-mocap calibration typically operates within a particular sensing pipeline, learning glove-signal-to-pose mappings, correcting IMU alignment and drift, or personalizing sensor-to-pose models \citep{wang2013glovecalib,li2025fsglove,zuo2025transformer,gosala2023selfcalibrated}. We instead formulate few-shot calibration at the derived 3D hand-keypoint interface, after a tracker has produced its pose estimate. Given \(K\) paired raw and target poses from a user and session, \MethodName{} conditions a HyperNet on those examples and generates LoRA updates for a frozen MANO-based estimator in one pass, without per-session gradient optimization. This couples a sensor-independent pose interface with amortized, session-conditioned weight generation: unlike a LoRA initialization shared across sessions, the generated updates depend on the current calibration set (Appendix~\ref{Appendix:LearnedPriorControls}). The formulation requires no access to device-specific raw signals; our real-device evaluation remains limited to one tensile glove.

Our contributions are threefold. First, we formulate calibration of derived 3D hand keypoints from a few pose pairs, separating the correction from device-specific sensor-to-pose calibration. Second, we generate session-conditioned LoRA weights for a frozen MANO estimator in one pass; learned-prior and set-conditioned residual controls test whether the gains stem from pair-conditioned weight generation beyond a shared prior or set conditioning alone. Third, we evaluate accuracy and adaptation cost on augmented InterHand2.6M and paired real-glove/AVP data, then test the resulting hand poses in physical teleoperation.

\section{Related Work}
\label{sec:related_work}

\textbf{Mocap Devices:}
High-fidelity dexterous teleoperation depends on reliable capture of the operator hand, yet existing mocap interfaces exhibit a strong accuracy-accessibility trade-off. Glove- and exoskeleton-based systems provide stable kinematics but are costly, cumbersome, and sensitive to user-specific geometry and donning conditions \citep{dipietro2008glove,wang2013glovecalib,zhang2025exocalib,li2025fsglove}. The resulting joint biases or fingertip drift propagate directly into the collected demonstrations. Markerless and RGB-based pipelines, including FreiHAND, MediaPipe Hands, and HaMeR, have improved accessibility \citep{romero2017mano,zimmermann2017rgbhand,zimmermann2019freihand,hampali2020honnotate,zhang2020mediapipe,moon2020interhand,Pavlakos_2024_CVPR}, but still struggle with occlusion, hand-object interaction, cross-user morphology, and temporal stability. \MethodName{} is complementary: rather than introducing a new tracker, it calibrates existing glove-based streams so that inexpensive devices become more consistent for downstream use.

\textbf{Prior Hand-Mocap Calibration:}
Wang and Neff~\citep{wang2013glovecalib} learn a mapping from glove measurements to hand motion. Gosala et al.~\citep{gosala2023selfcalibrated} use optical reference data to personalize a wearable sensor-to-pose model through side-tuning. FSGlove~\citep{li2025fsglove} jointly optimizes IMU alignment, hand shape, and kinematics, while the Transformer IMU Calibrator~\citep{zuo2025transformer} estimates time-varying IMU drift and offsets. These approaches address calibration within particular sensing pipelines. \MethodName{} instead takes their downstream 3D-keypoint representation as its input and learns to generate a session-specific correction of a frozen MANO estimator from a few paired poses. This interface makes the formulation independent of raw sensor modality; generalization beyond the evaluated tensile glove remains to be tested.

\textbf{Dexterous Hand Retargeting and Teleoperation:}
Recent teleoperation systems have made dexterous data collection cheaper, more immersive, and more hardware-agnostic. Vision-based systems such as DexPilot and AnyTeleop reduce dependence on instrumented gloves \citep{handa2019dexpilot,qin2023anyteleop,lin2024learningvisuotactileskillsmultifingered}, while VR-based platforms such as Open-TeleVision and OPEN TEACH improve operator immersion \citep{cheng2024opentelevision,iyer2024openteach}. In parallel, learning-based retargeting methods close the embodiment gap beyond purely kinematic alignment \citep{sivakumar2022telekinesis,park2025transferhandskills,li2025maniptrans,yin2025geometricretargetingprincipledultrafast}, and new master-side or robot-side designs improve transferability and contact richness \citep{wan2025rapidhand,fang2025dexop,junge2025adaptteleop}. 
These efforts largely assume that the upstream hand observation is already accurate; \MethodName{} targets that bottleneck and is compatible with the downstream stacks above.

\textbf{HyperNetworks and Parameter-Efficient Adaptation:}
HyperNetworks generate the weights of one network with another \citep{ha2017hypernetworks}, and task-conditioned variants support efficient cross-task adaptation \citep{vonoswald2020hypercl,mahabadi2021sharedhyper,zhou2025hyper,ren2025hypogen,sun2025hypertasrhypernetworkdriventaskawarescene}. In parallel, parameter-efficient methods such as Compacter and LoRA specialize a frozen backbone with compact low-rank updates \citep{mahabadi2021compacter,hu2022lora}, and hypernetworks have been shown to predict such updates from limited conditioning input \citep{ruiz2023hyperdreambooth}. 
\MethodName{} brings these ideas to dexterous teleoperation: \(h_{\psi}\) predicts LoRA-style updates \(\Delta W\) for the hand estimator from a handful of paired calibration poses, yielding fast and robust adaptation.

\section{Method}
\label{sec:method}

\subsection{Problem Formulation}
\label{sec:problem_formulation}
We consider the problem of calibrating noisy mocap hand poses using a small set of paired calibration examples. A hand observation is represented as 21 3D joint keypoints \(x\in\mathbb{R}^{21\times3}\). Let
\begin{equation}
    \mathcal{S}=\{(x_i^{n}, x_i^{c})\}_{i=1}^{K}
\end{equation}
denote the calibration set, where \(x_i^{n}\) is a noisy mocap hand-pose keypoint observation sampled from the noisy pose distribution \(\mathcal{D}_{\mathrm{noisy}}\), and \(x_i^{c}\) is its corresponding target 3D keypoint pose. The calibration set \(\mathcal{S}\) is assumed to capture the approximately stationary bias of the current setting, such as a particular subject, glove fit, or recording session.

We formulate calibration as the composition of two networks with distinct roles.

\textbf{Target network.} A MANO estimator
\begin{equation}
    g_{\phi}:\;\mathbb{R}^{21\times3}\longrightarrow\mathbb{R}^{45},
    \qquad
    \hat\theta = g_{\phi}(x),
\end{equation}
maps 21 3D hand keypoints to MANO pose parameters \(\hat\theta\). Its parameters \(\phi\) are trained once on clean data and then held fixed; per-user and per-session calibration acts on \(g_{\phi}\) only through additive LoRA updates rather than by re-optimizing \(\phi\).

\textbf{Hypernetwork.} A calibration HyperNet
\begin{equation}
    h_{\psi}:\;\bigl(\mathbb{R}^{21\times3}\times\mathbb{R}^{21\times3}\bigr)^{K}\longrightarrow\mathcal{W},
    \qquad
    \Delta W = h_{\psi}(\mathcal{S}),
\end{equation}
takes the calibration set as input and outputs a set of low-rank LoRA matrices \(\Delta W\), where \(\mathcal{W}\) denotes the space of LoRA adapters attached to \(g_{\phi}\). The parameters \(\psi\) are shared across all users and devices and constitute the only learned parameters of the calibration system; \(\phi\) is fixed.

\textbf{Calibrated estimator.} Injecting \(\Delta W\) into the target network yields a session-specific calibrated estimator
\begin{equation}
    g_{\phi + \Delta W}:\;\mathbb{R}^{21\times3}\longrightarrow\mathbb{R}^{45},
\end{equation}
which, composed with MANO forward kinematics, produces the calibrated 3D joints
\begin{equation}
    \hat x_{q}^{c} \;=\; \Pi_{\mathrm{MANO}}\!\bigl(g_{\phi + h_{\psi}(\mathcal{S})}(x_{q}^{n}),\,\beta\bigr)
    \;\eqqcolon\; f_{\psi}(x_{q}^{n};\,\mathcal{S}),
\end{equation}
for any new query \(x_{q}^{n}\sim\mathcal{D}_{\mathrm{noisy}}\). The shape parameter \(\beta\) is handled consistently across methods in each evaluation setting, as described in Section~\ref{sec:experiment}.

\textbf{Learning objective.} Since \(\phi\) is frozen, only the hypernetwork parameters \(\psi\) are trained, by minimizing the expected joint-position error of the resulting calibrated estimator:
\begin{equation}
    \min_{\psi}\;
    \mathbb{E}_{\mathcal{S},\,(x_{q}^{n}, x_{q}^{c})}
    \!\left[\,
        \mathcal{L}\!\bigl(f_{\psi}(x_{q}^{n};\,\mathcal{S}),\,x_{q}^{c}\bigr)
    \,\right],
\end{equation}
where \(\mathcal{L}\) is an \(L_{2}\) distance between hand joint positions. This objective encourages \(h_{\psi}\) to infer the underlying mocap bias from the paired calibration set and produce LoRA updates that correct unseen noisy poses drawn from the same distribution.

\begin{figure}[!t]
    \centering
    \includegraphics[width=0.98\linewidth]{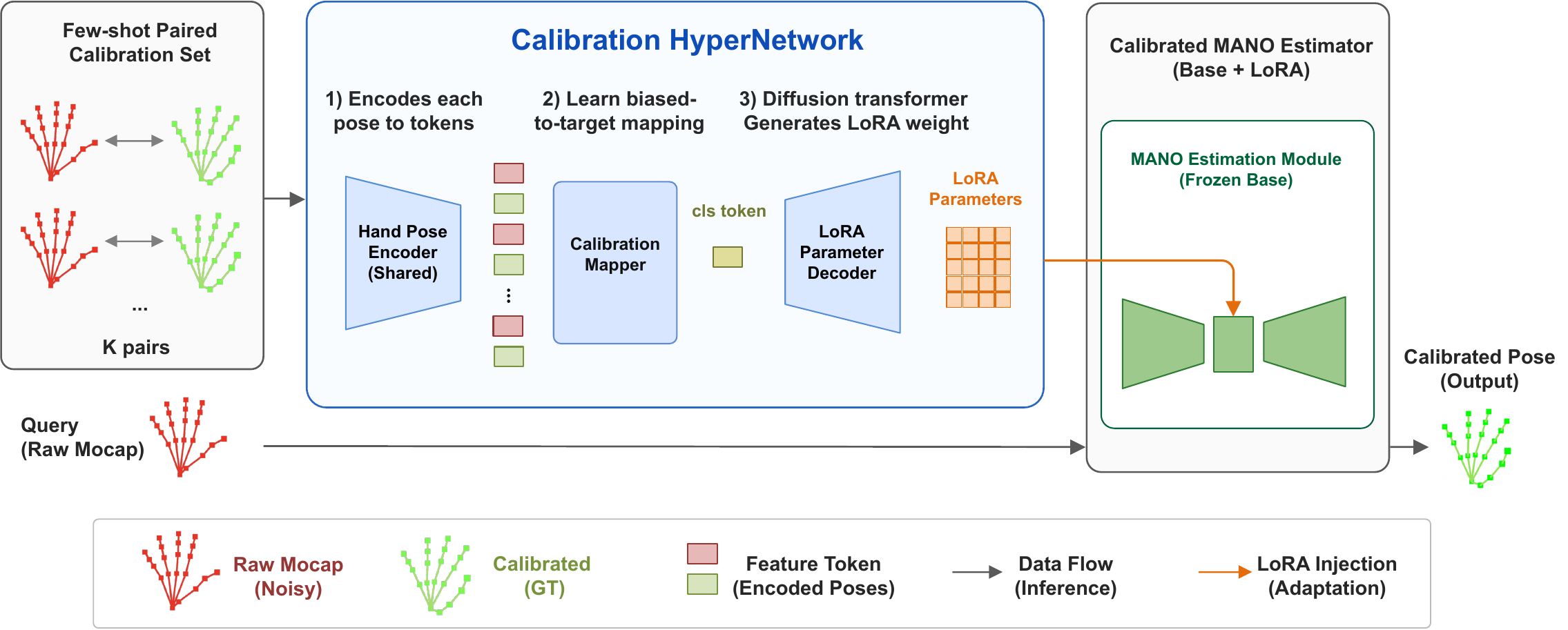}
    \caption{\MethodName{} model architecture.
    Given a calibration set \(\mathcal{S}=\{(x_i^{n},x_i^{c})\}_{i=1}^{K}\) of \(K\) paired noisy and target poses (left), the calibration HyperNet \(h_{\psi}\) (middle) encodes each pair, aggregates them through a transformer-based calibration mapper with one CLS token per LoRA matrix, and decodes the tokens into LoRA updates \(\Delta W\).
    These updates are injected once per session into the frozen MANO estimator \(g_{\phi}\) (right) to form \(g_{\phi+\Delta W}\), which maps each raw mocap query \(x_q^{n}\) at frame rate to calibrated MANO pose and 3D joints.
    }
    \label{fig:model_architecture}
\end{figure}

\subsection{Model Overview}
\label{sec:model_overview}
Figure~\ref{fig:model_architecture} illustrates how the target network \(g_{\phi}\) and the hypernetwork \(h_{\psi}\) defined in Section~\ref{sec:problem_formulation} are combined at deployment. Calibration consists of a one-time feed-forward conditioning pass followed by a frame-rate query path, both fully gradient-free:
(i) the HyperNet ingests the calibration set and outputs LoRA matrices, \(\Delta W = h_{\psi}(\mathcal{S})\);
(ii) these matrices are injected into the frozen MANO estimator to form the calibrated estimator \(g_{\phi + \Delta W}\);
(iii) each subsequent raw mocap pose \(x_{q}^{n}\) is mapped through \(g_{\phi + \Delta W}\) to obtain \(\hat\theta_{q}\), from which MANO reconstructs the calibrated joints \(\hat x_{q}^{c}\).
Step (i) is executed only once per session, while step (iii) runs at the mocap frame rate. The remainder of this section specifies the architectures of \(g_{\phi}\) (Section~\ref{sec:mano_arch}) and \(h_{\psi}\) (Section~\ref{sec:hyper_arch}).

\subsection{Target Network: MANO Estimator Architecture}
\label{sec:mano_arch}
The MANO estimator \(g_{\phi}\) consists of a hand-keypoint encoder and a MANO parameter decoder. The hand-keypoint encoder uses finger-wise CNN keypoint encoders followed by a transformer fusion layer. The MANO parameter decoder uses finger-wise MLPs and a unified MANO \(\theta\) decoder. The finger-wise design prevents each finger estimate from being overly affected by unrelated fingers, while the fusion layer captures global hand-pose correlation. The decoder is deliberately lightweight so that it can run in real time and accept LoRA updates from the HyperNet.

\subsection{Hypernetwork: Calibration HyperNet Architecture}
\label{sec:hyper_arch}
The calibration HyperNet \(h_{\psi}\) realizes the mapping \(\mathcal{S}\mapsto\Delta W\): its input is the \(K\) paired noisy/target poses in the calibration set, and its output is the LoRA low-rank matrices \(\Delta W\) injected into the frozen MANO estimator \(g_{\phi}\). To produce useful adapters from only a handful of pairs, \(h_{\psi}\) must infer the session-specific drift pattern from a small amount of runtime calibration data.

The HyperNet consists of three submodules: (1) hand-keypoint encoder, (2) calibration mapping, and (3) LoRA parameter decoder. The hand-keypoint encoder reuses the encoder from the MANO estimator \(g_{\phi}\) to extract pose features from each raw/calibrated pair. The calibration mapping is a transformer block with RoPE positional embedding \citep{su2023roformerenhancedtransformerrotary} that models the mapping relationship across the paired examples. It includes one learnable CLS token for each LoRA matrix to be generated; after calibration-map aggregation, each CLS token conditions the generation of its corresponding LoRA matrix. The LoRA parameter decoder is a DiT-style decoder \citep{peebles2023scalablediffusionmodelstransformers} that maps these LoRA-specific CLS tokens to the low-rank matrices in \(\Delta W\). We use the same LoRA adapter footprint for \MethodName{} and the LoRA-Finetune baseline; details are provided in Appendix~\ref{Appendix:LoRA-Details}.

\section{Model Training}
\label{sec:training}
Training proceeds in two stages that mirror the role separation introduced in Section~\ref{sec:problem_formulation}: first the target network \(g_{\phi}\) is trained on clean data, then it is frozen and the hypernetwork \(h_{\psi}\) is trained to generate LoRA updates that correct simulated mocap bias.

\subsection{Target Network Pretraining}
\label{sec:target_pretraining}
We train the parameters \(\phi\) of the MANO estimator \(g_{\phi}\) on clean InterHand2.6M data, using the mean squared error between the predicted MANO pose parameter \(\hat\theta = g_{\phi}(x)\) and the ground-truth MANO pose parameter. This pretraining establishes a hand-pose prior in MANO space that the HyperNet can later specialize through LoRA updates. After this stage, \(\phi\) is frozen and never updated again.

\subsection{Hypernetwork Training}
\label{sec:hyper_training}
With \(\phi\) fixed, we train the hypernetwork parameters \(\psi\) so that \(h_{\psi}\) produces LoRA updates \(\Delta W\) that compensate for realistic mocap bias. To expose \(h_{\psi}\) to a diverse family of bias patterns, we apply synthetic augmentations to InterHand2.6M sequences that simulate the drift behavior of glove-based mocap devices. The augmentations are designed to be strong and varied so that the augmented distribution covers, or at least overlaps with, realistic distribution shifts observed in real mocap data; the full strategy is provided in Appendix~\ref{Appendix:Augmentation Strategy}.

Each training iteration uses one augmented sequence as a session. From this sequence we sample \(K\) paired calibration examples \(\mathcal{S}=\{(x_i^{n}, x_i^{c})\}_{i=1}^{K}\) for the HyperNet, and a disjoint set of query frames \(\{(x_{q}^{n}, x_{q}^{c})\}\) for supervision. The HyperNet generates session-specific LoRA matrices \(\Delta W = h_{\psi}(\mathcal{S})\), which are injected into the frozen target network to form \(g_{\phi+\Delta W}\). Each query pose is then passed through this calibrated estimator and MANO forward kinematics to obtain \(\hat x_{q}^{c} = f_{\psi}(x_{q}^{n};\mathcal{S})\), and \(\psi\) is updated by minimizing the mean squared error between \(\hat x_{q}^{c}\) and the ground-truth calibrated joints \(x_{q}^{c}\). Gradients flow only through \(\psi\), and LoRA injection is differentiable in \(\Delta W\).

\section{Experiment}
\label{sec:experiment}
Our experiments are designed to evaluate \MethodName{} as a deployable calibration system rather than only as a pose-estimation model. We organize the evaluation around five research questions that together cover calibration accuracy, the source of \MethodName{}'s advantage, sample efficiency in \(K\), transfer to real low-cost glove data, and downstream impact on dexterous teleoperation:
(1) Does \MethodName{} reduce hand-pose error on biased mocap streams relative to uncalibrated input?
(2) How does \MethodName{} compare with optimization- and fine-tuning-based calibration baselines in accuracy and adaptation cost, and in particular, does the feed-forward HyperNet outperform gradient-based LoRA at matched adapter capacity?
(3) How does performance scale with the number of paired calibration poses \(K\)?
(4) Does \MethodName{} transfer from synthetic drift to real low-cost glove data across users and glove fits?
(5) Does improved hand-state calibration translate into measurable gains on downstream dexterous teleoperation tasks?

\subsection{Experiment Setup}
\textbf{Training datasets.} We pretrain the target network \(g_{\phi}\) on the training subset of InterHand2.6M, which provides a large pool of clean MANO-annotated hand poses captured by a high-precision multi-view mocap system. We then train the hypernetwork \(h_{\psi}\) on a synthetic dataset constructed by applying the structured drift augmentations described in Appendix~\ref{Appendix:Augmentation Strategy} to InterHand2.6M, while keeping \(\phi\) frozen. Each augmented sequence simulates one user/device/session with an internally consistent bias pattern, providing the diverse session-level distribution that the HyperNet must learn to summarize from only \(K\) paired examples.

\textbf{Test datasets.} We evaluate \MethodName{} on three settings of increasing realism. On the synthetic side, we apply the same augmentation family to the InterHand2.6M validation set, instantiating 20 augmented sets with distinct drift patterns; all reported synthetic numbers are aggregated over these 20 sets to assess robustness across drift types rather than performance on a single bias. For real low-cost-glove evaluation, we collect a paired dataset using an Apple Vision Pro (AVP) as a high-end reference tracker and a low-cost tensile glove worn simultaneously, treating the AVP 3D keypoints as the calibration target. Split and sequence statistics are provided in Appendix~\ref{Appendix:RealWorldData}. For downstream evaluation, we deploy \MethodName{} in a live teleoperation system on four physical manipulation tasks.

\textbf{Evaluation setup.} Unless otherwise specified, all methods use \(K=4\) paired calibration poses. For offline evaluations on InterHand2.6M and the real glove dataset, \(\mathcal{S}\) is sampled from the first 200 frames of each validation sequence with a fixed random seed; the remaining frames form the evaluation split. The real-glove targets are synchronized AVP reference poses. For deployment, we record a fixed library of four 21-joint target poses with AVP once offline: index pinch, fully open hand, claw/half-close, and four straight fingers with the thumb near the palm (Appendix~\ref{Appendix:RealWorldData}). The targets use the mocap stream's joint order and coordinate convention. At the start of each teleoperation trial, the operator performs these prompts wearing only the low-cost glove; each raw observation is paired with its stored library target. No AVP is needed at runtime. All methods share the same shape-handling protocol and do not calibrate MANO shape \(\beta\); details are in Appendices~\ref{Appendix:Mano Beta Handling} and~\ref{Appendix:Target Pose Selection}.

\subsubsection{Teleoperation Protocol}
\label{sec:teleop_protocol}
We evaluate four physical tasks with a right robot hand: removing a fixed bottle's cap by twisting; lifting an earbud with a thumb--middle-finger pinch and holding it for 3~s; lifting a small glue bottle with the thumb, index, and middle fingers and holding it for 3~s; and unlocking a tissue-box lid by stabilizing it with four fingers and pulling with the thumb. These outcomes define task success. Two operators participate: one performs bottle-cap and earbud tasks, and the other performs glue-bottle and tissue-box tasks. The same operator tests all three calibration conditions for each assigned task. Operators can directly see the robot and thus can compensate for control error, but are blinded to the calibration method's identity. For each method--task pair, they receive 5~min of familiarization and then perform 10 evaluation trials; method order is randomized within each task. All conditions use the same trial budget and the same geometric-retargeting method~\citep{yin2025geometricretargetingprincipledultrafast}. A separate retargeting model is trained for each calibration condition using the same configuration. We report success counts and completion time, with the failure penalty for overall time specified in Table~\ref{tab:downstream_dexterous}.

\textbf{Evaluation metrics.} We report a comprehensive set of pose, geometry, and deployment metrics. MPJPE and PA-MPJPE measure 3D joint accuracy before and after Procrustes alignment: MPJPE preserves residual offset, scale, and orientation bias, while PA-MPJPE isolates articulated pose accuracy. AUCJ is the normalized area under the joint PCK curve, computed over 101 uniformly spaced thresholds from 0 to 0.05~m. Tip-PA is the PA-aligned mean L2 error over the five MANO fingertips and is reported separately because fingertip geometry dominates dexterous-manipulation outcomes. Pinch error is the L2 distance between two fingertips on frames where the ground-truth pinch distance is below 10~mm; this isolates the regime that contact-rich tasks depend on. Calibration time is the test-time conditioning or adaptation cost incurred before applying the calibrated estimator: for \MethodName{}, this is one forward pass of \(h_{\psi}\); for fine-tuning baselines, it is gradient-based adaptation on the paired calibration set. For downstream evaluation, we report task success count over 10 trials and average completion time.

\subsection{Experiment Results}

\begin{table*}[t]
    \centering
    \vspace{-0.5em}
    \footnotesize
    \setlength{\tabcolsep}{2.5pt}
    \renewcommand{\arraystretch}{1.12}
    \setlength{\extrarowheight}{1pt}
    \begin{tabular}{@{}lcccccc@{}}
        \toprule
        Method 
        & MPJPE$\downarrow$ 
        & PA-MPJPE$\downarrow$ 
        & Tip-PA$\downarrow$ 
        & AUCJ$\uparrow$ 
        & Pinch$\downarrow$ 
        & Calib. s$\downarrow$ \\
        & (mm) 
        & (mm) 
        & (mm) 
        & 
        & (mm) 
        & (s) \\
        \midrule
        No calibration 
        & 21.52 & 16.55 & 37.15 & 0.611 & 48.49 & 0.00 \\
        \midrule
        Ridge 
        & 17.62 & 13.00 & 28.28 & 0.676 & 45.35 & \textbf{1.6} \\
        Last-layer-Finetune 
        & 16.62 & 12.01 & 25.92 & 0.692 & 39.18 & 45.6 \\
        Full-Finetune 
        & 18.11 & 12.77 & 29.21 & 0.668 & 37.11 & 95.4 \\
        LoRA-Finetune 
        & 16.38 & 11.67 & 25.33 & 0.694 & 37.28 & 89.4 \\
        \textbf{\MethodName{} (ours)} 
        & \textbf{15.64} 
        & \textbf{10.45} 
        & \textbf{23.29} 
        & \textbf{0.702} 
        & \textbf{34.90} 
        & 11.2 \\
        \bottomrule
    \end{tabular}
    \caption{Calibration results on the augmented InterHand2.6M evaluation setting. Calib. s reports the total adaptation time in seconds for whole validation sets on one H100.}
    \label{tab:interhand}
    \vspace{-0.5em}
\end{table*}

\textbf{RQ1: \MethodName{} substantially reduces hand-pose error on biased mocap streams.}
On the augmented InterHand2.6M evaluation set, \MethodName{} reduces PA-MPJPE from 16.55~mm (no calibration) to 10.45~mm, a 37\% relative reduction, with corresponding gains across MPJPE, Tip-PA, AUCJ, and pinch error (Table~\ref{tab:interhand}). The improvement is largest on fingertip-sensitive metrics (Tip-PA drops from 37.15 to 23.29~mm), which matches the design intuition that calibration matters most where small biases reshape contact geometry. These results confirm that the HyperNet successfully infers session-level drift from only \(K=4\) paired examples and that the predicted LoRA updates \(\Delta W\) correct unseen queries from the same session.

\textbf{RQ2: \MethodName{} outperforms optimization- and fine-tuning-based baselines, and the advantage persists at matched LoRA capacity.}
We compare against five baselines spanning the natural spectrum of calibration strategies: (1) \emph{No Calibration}, which uses raw mocap output; (2) \emph{Ridge}, an optimization-based linear mapping fitted in MANO \(\theta\) space from the paired examples; (3) \emph{Last-layer-Finetune}, which gradient-tunes only the final MLP of the MANO \(\theta\) decoder on the paired examples; (4) \emph{Full-Finetune}, which gradient-tunes the entire \(g_{\phi}\); and (5) \emph{LoRA-Finetune}, which injects the same LoRA adapter set as \MethodName{} into \(g_{\phi}\) and tunes only those adapters by gradient descent on the paired examples. LoRA-Finetune matches the adapter footprint, but its cold start does not control for \MethodName{}'s learned cross-session prior. Appendix~\ref{Appendix:LearnedPriorControls} adds a meta-learned LoRA initialization, a set-conditioned residual predictor, and a published side-tuning baseline to examine that distinction.

As shown in Table~\ref{tab:interhand}, \MethodName{} achieves the best score on every precision metric while being substantially faster than every gradient-based baseline: 11.2~s of aggregate H100 conditioning for \MethodName{} versus 45.6--95.4~s for the fine-tuning baselines. The accuracy advantage over LoRA-Finetune at matched adapter capacity (PA-MPJPE 10.45 vs.~11.67~mm, Tip-PA 23.29 vs.~25.33~mm) shows that the gains are not attributable to the LoRA adapters alone: feed-forward weight generation by \(h_{\psi}\) outperforms per-session gradient descent over the same parameter space. Ridge regression is the only baseline competitive on adaptation time, but its accuracy degrades severely. The deployment-latency gap is even larger on CPU, where \MethodName{} requires 13.16~s versus 34.47--223.25~s for gradient fine-tuning baselines (Appendix~\ref{Appendix:CPUCalibrationTime}); this is the regime that matters for in-the-field teleoperation rigs without a dedicated GPU, which signifies the applicability of \MethodName{}.

\Needspace{0.35\textheight}
\begin{wrapfigure}{r}{0.50\textwidth}
    \centering
    \vspace{-0.5\baselineskip}
    \includegraphics[width=\linewidth]{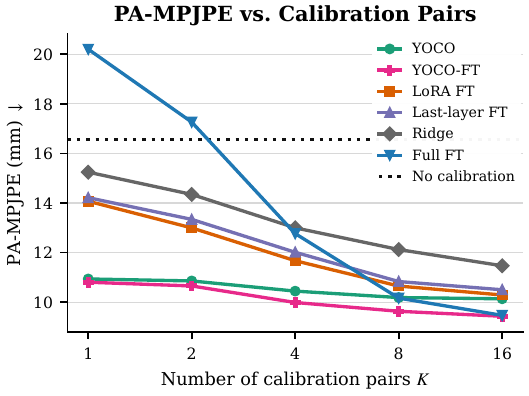}
    \caption{PA-MPJPE versus calibration-pair count, including \MethodName{} followed by test-time fine-tuning (YOCO-FT). The dotted line marks no calibration.}
    \label{fig:interhand_num_pairs_pa_mpjpe}
    \vspace{-0.5\baselineskip}
\end{wrapfigure}

\textbf{RQ3: \MethodName{} provides strong one-pass few-shot calibration.}
Figure~\ref{fig:interhand_num_pairs_pa_mpjpe} sweeps \(K\) from 1 to 16. Among methods that calibrate without further test-time optimization, \MethodName{} achieves the lowest PA-MPJPE at \(K\in\{1,2,4\}\): collecting four paired poses takes a few seconds, whereas collecting sixteen is closer to a minute. At \(K=1\), \MethodName{} reaches 10.94~mm, below every gradient-based baseline at \(K=4\). At \(K=8\), gradient-based baselines close the gap, and at \(K=16\), Full-Finetune slightly surpasses the one-pass \MethodName{}. The new YOCO-FT curve shows the effect of adding per-session fine-tuning after \MethodName{} initialization; it improves accuracy further but gives up gradient-free adaptation. This pattern is consistent with an optimization or capacity limit at larger \(K\), but does not isolate which factor dominates. The numerical multi-method sweep is reported in Appendix~\ref{Appendix:K-Sweep}.

\Needspace{0.30\textheight}
\begin{wrapfigure}{r}{0.50\linewidth}
    \centering
    \vspace{-0.5\baselineskip}
    \includegraphics[width=\linewidth]{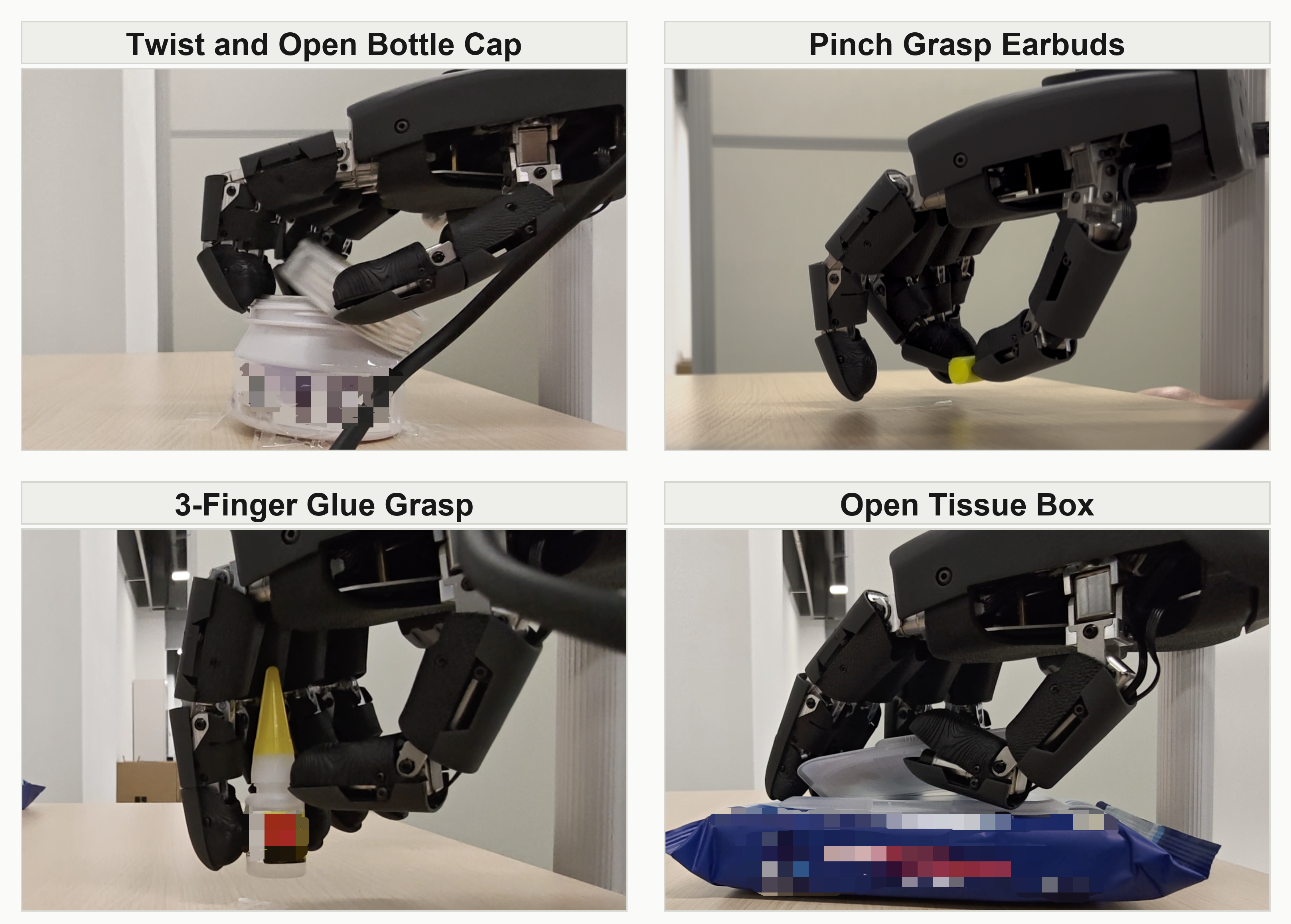}
    \caption{Goal states for the downstream dexterous manipulation tasks: twisting and opening a bottle cap, pinch-grasping earbuds, three-finger glue grasping, and tissue-box opening.}
    \label{fig:task_setting}
    \vspace{-0.5\baselineskip}
\end{wrapfigure}

\textbf{RQ4: \MethodName{} transfers to real low-cost glove data and generalizes across users.}
We evaluate on the paired AVP/tensile-glove dataset described in Appendix~\ref{Appendix:RealWorldData}, which exposes \MethodName{} to the actual user- and fit-dependent distortions of a low-cost glove rather than synthetic drift. As shown in Table~\ref{tab:real_mocap}, \MethodName{} attains the best score on every primary metric (MPJPE, PA-MPJPE, Tip-PA, AUCJ, pinch error) while requiring 6.73~s of conditioning versus 14.57~s for LoRA-Finetune and 13.04~s for Full-Finetune. The per-user standard deviations in Appendix~\ref{Appendix:RealWorldUserStd} indicate that this advantage is consistent across the four users in the dataset rather than driven by any single operator. Notably, Ridge regression degrades below the no-calibration baseline on real data, which suggests that the bias structure of a real glove is non-linear enough that linear correction in MANO \(\theta\) space is actively harmful; this is the regime where a learned, structured calibrator earns its keep.

\begin{table*}[t]
    \centering
    \vspace{-0.5em}
    \footnotesize
    \setlength{\tabcolsep}{2pt}
    \renewcommand{\arraystretch}{0.95}
    \setlength{\extrarowheight}{1pt}
    \begin{tabular}{@{}lcccccc@{}}
        \toprule
        Method & MPJPE$\downarrow$ & PA-MPJPE$\downarrow$ & Tip-PA$\downarrow$ & AUCJ$\uparrow$ & Pinch$\downarrow$ & Calib.$\downarrow$ \\
        & (mm) & (mm) & (mm) & & (mm) & (s) \\
        \midrule
        No calibration 
        & 20.49 
        & 14.63 
        & 36.22 
        & 0.610 
        & 51.39 
        & 0.00 \\
        \midrule
        Ridge 
        & 24.71 
        & 19.21 
        & 45.50 
        & 0.573 
        & 57.54 
        & \textbf{1.49} \\

        Last-layer-Finetune 
        & 18.86 
        & 14.67 
        & 33.88 
        & 0.643 
        & 43.63 
        & 5.04 \\

        Full-Finetune 
        & 20.00 
        & 15.73 
        & 36.67 
        & 0.630 
        & 49.89 
        & 13.04 \\

        LoRA-Finetune 
        & 18.72 
        & 13.94 
        & 33.23 
        & 0.648 
        & 54.01 
        & 14.57 \\

        \textbf{\MethodName{} (ours)} 
        & \textbf{17.75} 
        & \textbf{13.14} 
        & \textbf{31.73} 
        & \textbf{0.657} 
        & \textbf{42.79} 
        & 6.73 \\
        \bottomrule
    \end{tabular}
    \caption{Real-world mocap calibration results. Metrics are averaged over 3 seeds. Calib. s reports the total conditioning or adaptation time in seconds for whole datasets on one H100. Corresponding user-level std is provided in Appendix~\ref{Appendix:RealWorldUserStd}.}
    \label{tab:real_mocap}
    \vspace{-0.5em}
\end{table*}

\textbf{RQ5: The \MethodName{} teleoperation pipeline improves task success.}
The tasks, operator assignment, visual feedback, familiarization, retargeting, and trial order are described in Section~\ref{sec:teleop_protocol}.

As shown in Table~\ref{tab:downstream_dexterous}, \MethodName{} achieves an overall success rate of 90\%, compared with 65\% for LoRA-Finetune and 35\% for the uncalibrated baseline, and the lowest overall completion time across the four tasks. The largest per-task gains appear on contact-sensitive tasks: Earbud Pinch Grasping rises from 1/10 (LoRA-Finetune) to 8/10 (\MethodName{}), and Tissue Box Opening from 8/10 to 10/10. The earbud task uses a middle-finger pinch, whereas the calibration poses include an index-finger pinch but no middle-finger pinch, suggesting transfer to an unobserved fingertip configuration. Because retargeting models are trained separately for each condition, the task results characterize the complete teleoperation pipelines rather than isolating calibration as the sole cause.

\begin{table*}[ht]
    \centering
    \setlength{\tabcolsep}{2.5pt}
    \renewcommand{\arraystretch}{1.12}
    \setlength{\extrarowheight}{1pt}
    \resizebox{\textwidth}{!}{%
    \begin{tabular}{@{}lcccccccccc@{}}
        \toprule
        Method 
        & \multicolumn{2}{c}{\shortstack{Bottle Cap\\[-0.3ex]Opening}} 
        & \multicolumn{2}{c}{\shortstack{Earbud Pinch\\Grasping}} 
        & \multicolumn{2}{c}{\shortstack{Glue Three-Finger\\[-0.3ex]Grasping}} 
        & \multicolumn{2}{c}{\shortstack{Tissue Box\\Opening}} 
        & \multicolumn{2}{c}{Overall} \\
        \addlinespace[1pt]
        & Success$\uparrow$ & Time (s)$\downarrow$ 
        & Success$\uparrow$ & Time (s)$\downarrow$ 
        & Success$\uparrow$ & Time (s)$\downarrow$ 
        & Success$\uparrow$ & Time (s)$\downarrow$ 
        & Success Rate$\uparrow$ & Time (s)$\downarrow$ \\
        \midrule
        No calibration 
        & 9/10 & 35.6 
        & 0/10 & -- 
        & 5/10 & 7.0 
        & 0/10 & -- 
        & 35\% & 24.4 \\

        LoRA-Finetune 
        & 9/10 & 34.1 
        & 1/10 & 6.0 
        & \textbf{8/10} & \textbf{6.6} 
        & 8/10 & \textbf{13.3} 
        & 65\% & 17.9 \\

        \textbf{\MethodName{} (ours)} 
        & \textbf{10/10} & \textbf{27.3} 
        & \textbf{8/10} & \textbf{4.2} 
        & \textbf{8/10} & 8.9 
        & \textbf{10/10} & 17.4 
        & \textbf{90\%} & \textbf{14.9} \\
        \bottomrule
    \end{tabular}%
    }
    \caption{Downstream dexterous teleoperation results. We evaluate 4 tasks and count each task success over 10 trials and average completion time over successful trials. Failed trials are capped at one minute and excluded from task-specific average completion time, while each failed trial is penalized by assigning it the longest observed trial time in the overall average time. More stats are provided in Appendix~\ref{Appendix:Additional Teleoperation Statistics}.}
    \label{tab:downstream_dexterous}
\end{table*}

\section{Conclusion}
\label{sec:conclusion}
\MethodName{} reframes mocap calibration as feed-forward weight generation: a hypernetwork \(h_{\psi}\) reads a handful of paired poses and emits LoRA updates \(\Delta W\) that specialize a frozen MANO estimator \(g_{\phi}\), avoiding per-session optimization of the estimator. The downstream results show promising teleoperation gains for the evaluated calibration-and-retargeting pipelines.

\paragraph{Limitations}
\MethodName{} focuses on pose calibration and leaves MANO shape estimation to a separate stage, which integrates cleanly with existing shape-fitting pipelines. The session-stationary bias assumption is consistent with sensor manufacturing tolerances of most commercial gloves, but high-frequency drift on less stable hardware would require re-calibration or a temporal extension of the HyperNet. The augmentation pipeline lets us scale training across diverse drift patterns at low cost, though covering high-frequency drift remains an open direction tied to sensor characterization.

\section*{Acknowledgments}

This work is supported by the Early Career Scheme of the Research Grants Council (RGC) grant \# 27207224, the HKU-100 Award, a donation from the Musketeers Foundation, and in part by the JC STEM Lab of Autonomous Intelligent Systems funded by The Hong Kong Jockey Club Charities Trust and computing resource provided by HKU-Shanghai Intelligent Computing Research Center. We thank TranscEngram for supporting data collection and providing hardware used in this project. This project is also supported and funded by Shenzhen Loop Area Institute under project FPF10120250005.

\clearpage
\bibliography{example}  %

\clearpage
\appendix
\section{Method Details}
\subsection{Augmentation Strategy for HyperNet training}
\label{Appendix:Augmentation Strategy}
To train HyperLoRA under realistic calibration errors in glove-based mocap devices, we generate synthetic
biases in the MANO pose space rather than directly perturbing 3D keypoints.
Let $\theta_{d,t}\in{R}^{45}$ denote the MANO hand pose of frame $t$ in
sequence $d$, reshaped as $\theta_{d,t}^{f,j}\in{R}^{3}$ for finger
$f$ and joint level $j$. For each sequence, we sample a bias style $\phi_d$
once and keep it fixed across all frames. This reflects the observation that
glove or hand-tracking miscalibration is usually session-consistent, rather
than independent frame-wise noise.

The sampled bias combines several physically plausible error modes:
per-finger zero offsets, per-joint residual offsets, global/finger/joint gain
errors, adjacent-finger crosstalk, and proximal-to-distal coupling within each
finger. Concretely, the biased pose is generated as
\begin{equation}
\tilde{\theta}_{d,t}^{f,j}
=
s_d^{f,j}\theta_{d,t}^{f,j}
+
a_{f,j}\sum_{g\in\mathcal{N}(f)}
C_d^{f,g}s_d^{g,j}\theta_{d,t}^{g,j}
+
\sum_{k<j} U_d^{j,k}s_d^{f,k}\theta_{d,t}^{f,k}
+
o_d^{f,j},
\end{equation}
where $s_d^{f,j}$ is a clipped multiplicative gain, $C_d$ models leakage
between adjacent fingers, $U_d$ models tendon-like proximal-to-distal coupling,
$o_d^{f,j}$ is the additive offset, and $a_{f,j}$ attenuates perturbations
from proximal to distal joints. These parameters are sampled from zero-mean
Gaussian families with Bernoulli activation probabilities, so each sequence
receives a different but internally consistent bias type.

We then apply MANO forward kinematics to both the calibrated and noisy poses:
\begin{equation}
x_{d,t}^{c}=\Pi_{\mathrm{MANO}}(\theta_{d,t}),
\qquad
x_{d,t}^{n}=\Pi_{\mathrm{MANO}}(\tilde{\theta}_{d,t}),
\end{equation}
and store the resulting pair $(x_{d,t}^{n}, x_{d,t}^{c})$ offline. During
training, the calibration model receives the noisy joints $x_{d,t}^{n}$ as
input and predicts the calibrated MANO pose. \MethodName{} is conditioned by a small
set of calibration examples
\begin{equation}
\mathcal{P}_d
=
\left\{(x_{d,\tau_i}^{n},x_{d,\tau_i}^{c})\right\}_{i=1}^{K},
\end{equation}
which are encoded by the calibration mapping to generate
sequence-specific LoRA parameters. This augmentation exposes the model to a
diverse family of structured, kinematically valid calibration errors while
preserving the temporal consistency expected from real glove or hand-tracking
biases.

\subsection{LoRA Adaptation Details}
\label{Appendix:LoRA-Details}
All layers in the MANO estimator are equipped with LoRA adapters. Unless otherwise specified, both \MethodName{} and the LoRA-Finetune baseline use LoRA rank \(r=8\) and scaling factor \(\alpha=8\). \MethodName{} generates the low-rank matrices for this full adapter set from the HyperNet, while the LoRA-Finetune baseline optimizes the same adapter set by gradient descent on the paired calibration examples. This keeps the adaptation capacity fixed across the feed-forward and optimization-based LoRA comparisons.

\Needspace{0.7\textheight}
\section{Learned-Prior Controls and Additional Baselines}
\label{Appendix:LearnedPriorControls}
Cold-start LoRA-Finetune uses the same adapter footprint as \MethodName{}, but lacks its learned prior across calibration sessions. To separate this effect from conditioning on the specific \(K\) pairs, \emph{Meta-LoRA} learns a session-independent LoRA initialization using the same target estimator and adapter footprint; \emph{Meta-LoRA+FT} adapts that initialization by gradient descent on the pairs. \emph{Set-Cond. Residual} instead predicts MANO-pose residuals from the paired set without test-time gradients. \emph{YOCO+LoRA-FT} fine-tunes the adapters generated by \MethodName{}. We also adapt the published side-tuning calibration approach of \citet{gosala2023selfcalibrated}; its original sensor inputs and supervision differ, so this is a related-method comparison rather than a matched-device calibration protocol.

\begin{table}[htbp]
    \centering
    \footnotesize
    \setlength{\tabcolsep}{4pt}
    \begin{tabular}{lcccc}
        \toprule
        & \multicolumn{2}{c}{Real glove (AVP reference)} & \multicolumn{2}{c}{Augmented InterHand2.6M} \\
        \cmidrule(lr){2-3}\cmidrule(lr){4-5}
        Method & PA-MPJPE$\downarrow$ & AUCJ$\uparrow$ & PA-MPJPE$\downarrow$ & AUCJ$\uparrow$ \\
        \midrule
        LoRA-Finetune & 13.94 & 0.648 & 11.67 & 0.694 \\
        Set-Cond. Residual & 13.74 & 0.645 & 12.66 & 0.664 \\
        Side-Tuning~\citep{gosala2023selfcalibrated} & 13.48 & 0.641 & 15.35 & 0.618 \\
        Meta-LoRA & 13.51 & 0.633 & 13.94 & 0.639 \\
        Meta-LoRA+FT & 14.17 & 0.664 & 11.41 & 0.702 \\
        \MethodName{} & \textbf{13.14} & 0.657 & 10.45 & 0.702 \\
        \MethodName{}+LoRA-FT & 13.28 & \textbf{0.665} & \textbf{9.99} & \textbf{0.707} \\
        \bottomrule
    \end{tabular}
    \caption{Learned-prior controls and additional calibration baselines at \(K=4\). PA-MPJPE is in millimeters; AUCJ is the normalized joint-PCK area.}
    \label{tab:learned_prior_controls}
\end{table}

Table~\ref{tab:learned_prior_controls} shows that, compared with the session-independent Meta-LoRA initialization, \MethodName{} reduces PA-MPJPE from 13.51 to 13.14~mm on the real glove data and from 13.94 to 10.45~mm on augmented InterHand2.6M. The set-conditioned residual control and Side-Tuning provide additional checks on the adaptation mechanism and prior work. Further LoRA fine-tuning improves some scores, at the cost of test-time optimization.

\section{Experiment Details}
\subsection{MANO Beta Handling}
\label{Appendix:Mano Beta Handling}
For the evaluation on InterHand2.6M dataset, we use the ground truth shape parameter for all methods.
For the evaluation on real-world dataset, we use optimization-based shape estimation method to estimate the shape parameter for all methods.
For the evaluation on downstream dexterous manipulation tasks, we use a MANO shape parameter \(\beta\) pre-computed when collecting the AVP mocap poses for the preset-pose library, and hold this same \(\beta\) fixed for all methods.

\subsection{Target Pose Selection}
\label{Appendix:Target Pose Selection}
For all offline evaluation, we report the average performance over 3 random seeds for the selection of paired calibration poses. 
In addition, the calibration poses are selected only from the first 200 frames of the validation sequence, and evaluation is performed on the remaining frames
to avoid data leakage and evaluate the generalization ability of \MethodName{} and the baselines to uncalibrated poses.

\subsection{Real-World Data Protocol}
\label{Appendix:RealWorldData}
Our real-world glove/mocap dataset contains four users. For each frame, the noisy observation \(x^n\) is the low-cost glove-derived hand keypoint pose, while the calibrated target \(x^c\) is the synchronized 3D hand keypoint pose from Apple Vision Pro (AVP). AVP is used here as a high-end reference source for offline evaluation data, not as a device required during deployment-time calibration. Each user contributes one sequence for evaluation. Each sequence is approximately 1500 frames long at 30 FPS. 

\begin{figure}[ht]
    \centering
    \footnotesize
    \begin{minipage}[t]{0.23\linewidth}
        \centering
        \includegraphics[width=\linewidth,height=0.16\textheight,keepaspectratio]{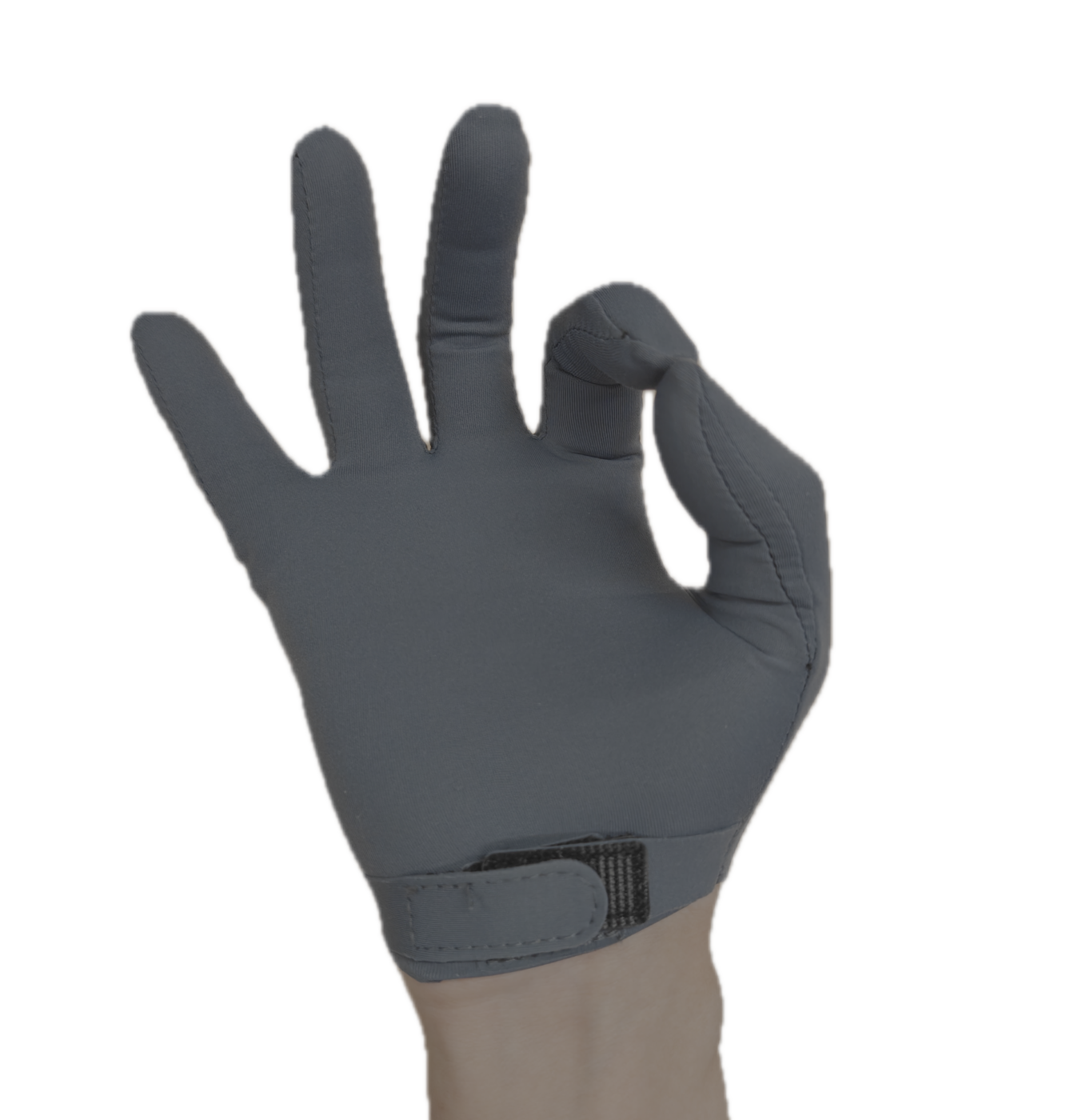}
        \vspace{0.3em}
        \textbf{Index pinch.} Thumb tip and index fingertip touch; remaining fingers stay naturally open.
    \end{minipage}
    \hfill
    \begin{minipage}[t]{0.23\linewidth}
        \centering
        \includegraphics[width=\linewidth,height=0.16\textheight,keepaspectratio]{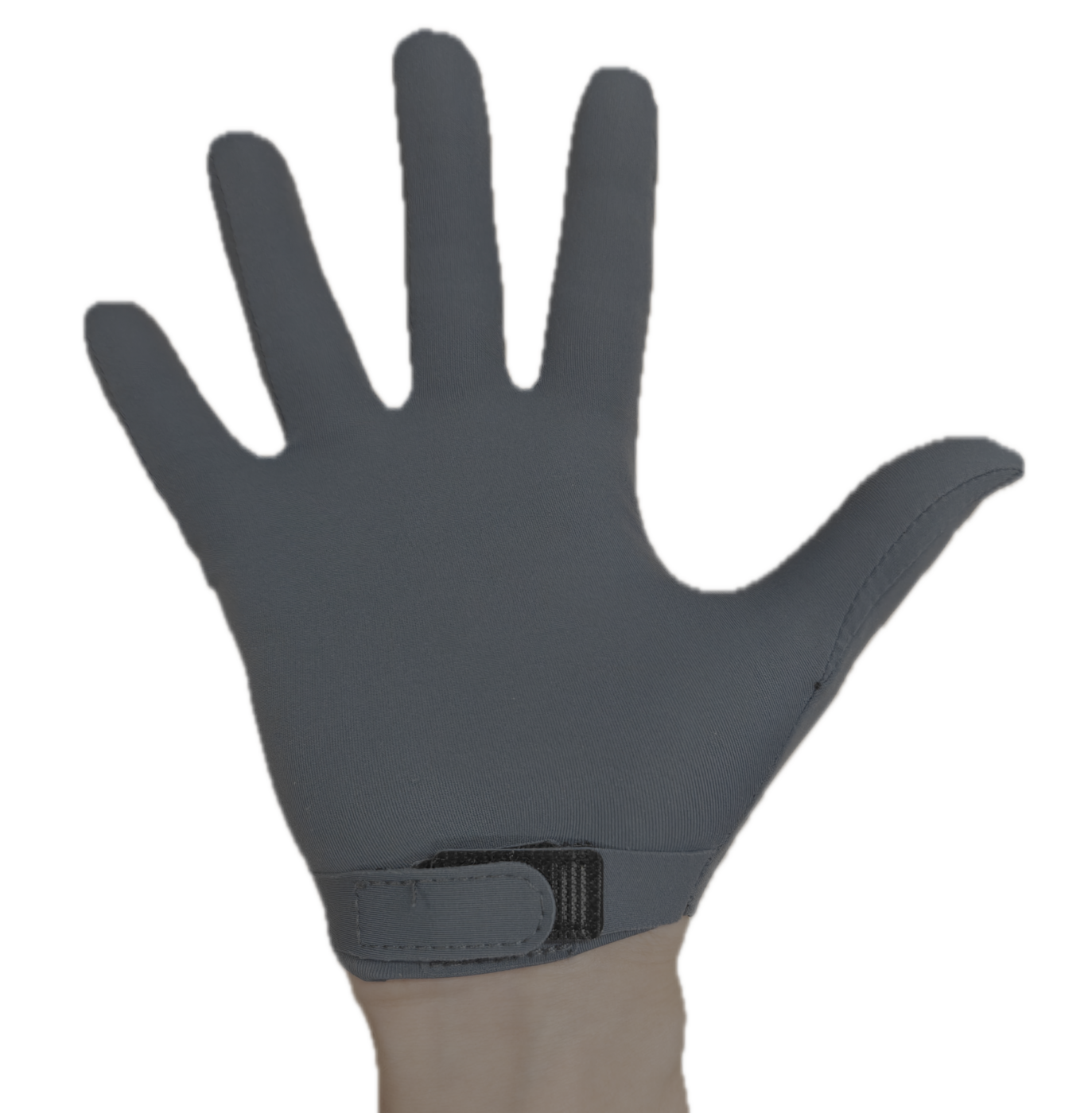}
        \vspace{0.3em}
        \textbf{Fully open.} Palm open with all five fingers extended.
    \end{minipage}
    \hfill
    \begin{minipage}[t]{0.23\linewidth}
        \centering
        \includegraphics[width=\linewidth,height=0.16\textheight,keepaspectratio]{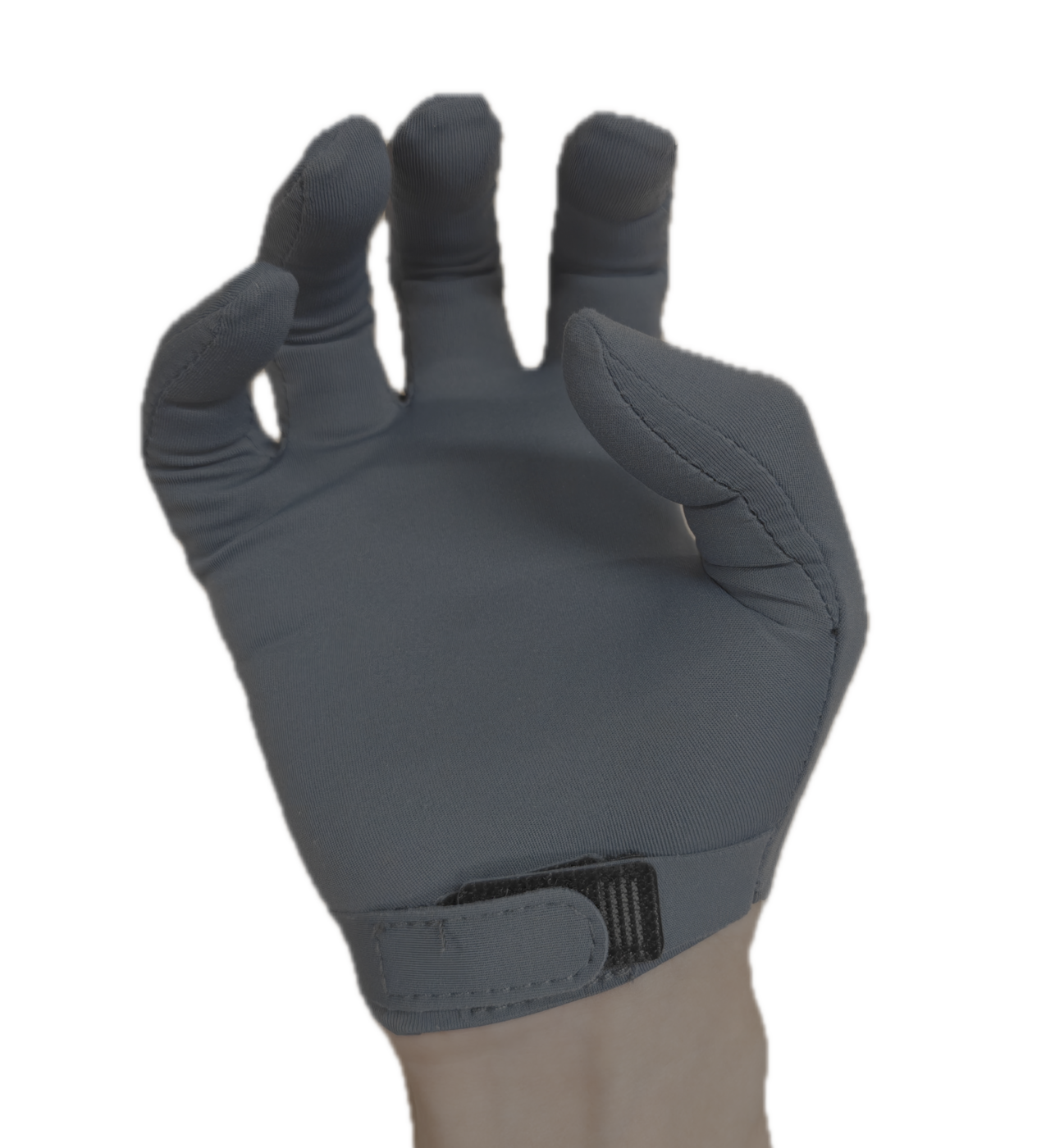}
        \vspace{0.3em}
        \textbf{Claw / half close.} Fingers are partially flexed in a mid-closure hand shape.
    \end{minipage}
    \hfill
    \begin{minipage}[t]{0.23\linewidth}
        \centering
        \includegraphics[width=\linewidth,height=0.16\textheight,keepaspectratio]{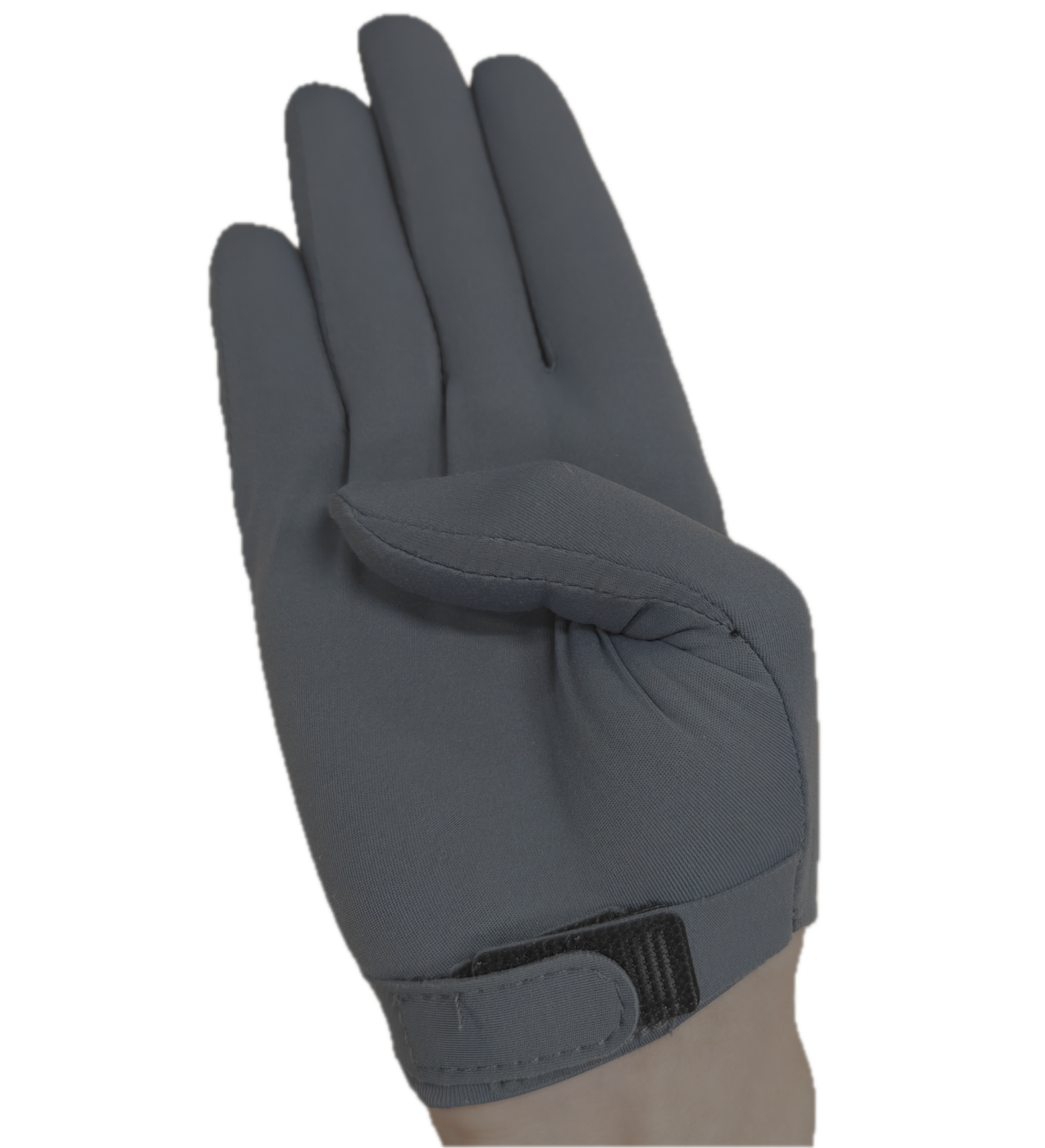}
        \vspace{0.3em}
        \textbf{Four fingers straight, thumb near palm.} Index, middle, ring, and little fingers are straight while the thumb is adducted close to the palm.
    \end{minipage}
    \caption{Preset pose library used to obtain deployable calibration targets. The four prompts are index pinch, fully open, claw / half close, and four fingers straight with the thumb near the palm.}
    \label{fig:deployment_pose_library}
\end{figure}

\subsection{CPU Deployment Calibration Time}
\label{Appendix:CPUCalibrationTime}
The Calibration time in Table~\ref{tab:interhand} and \ref{tab:real_mocap} are computed using H100 GPU. However, in practical teleoperation deployments, calibration may need to run on a CPU-only host. Table~\ref{tab:cpu_calibration_time} reports the average CPU adaptation time required before inference. We use the HyperNet forwarding time for \MethodName{} and the optimization/solving time for gradient-based or regression baselines.

\begin{table*}[ht]
    \centering
    \footnotesize
    \setlength{\tabcolsep}{2.5pt}
    \begin{tabular}{lcc}
        \toprule
        Method & Timing Source & CPU calibration time (s) \\
        \midrule
        \MethodName{} (ours) & \texttt{HyperNet Forward} & 13.16 \\
        Ridge & \texttt{Closed Form Solver} & 0.37 \\
        Last-layer-Finetune & \texttt{Gradient Descent} & 34.47 \\
        LoRA-Finetune & \texttt{Gradient Descent} & 59.58 \\
        Full-Finetune & \texttt{Gradient Descent} & 223.25 \\
        \bottomrule
    \end{tabular}
    \caption{Average CPU calibration time required before real deployment. }%
    \label{tab:cpu_calibration_time}
\end{table*}

\subsection{Real-World Per-User Variability}
\label{Appendix:RealWorldUserStd}
Table~\ref{tab:real_mocap_user_std} reports the per-user standard deviation of the real-world mocap calibration results. This table corresponds to Table~\ref{tab:real_mocap} to show how much each method varies across users.

\begin{table*}[ht]
    \centering
    \footnotesize
    \setlength{\tabcolsep}{3pt}
    \begin{tabular}{lcccccc}
        \toprule
        Method & MPJPE$\downarrow$ & PA-MPJPE$\downarrow$ & Tip-PA$\downarrow$ & AUCJ$\uparrow$ & Pinch$\downarrow$ & Calib. s$\downarrow$ \\
          & (mm) & (mm) & (mm) &  & (mm) & (s) \\
        \midrule
        No calibration & 1.14 & 0.76 & 3.18 & 0.021 & 7.01 & 0.00 \\
        \midrule
        Ridge & 4.32 & 2.95 & 9.12 & 0.053 & 12.53 & 0.68 \\
        Last-layer-Finetune & 0.87 & 1.01 & 2.47 & 0.017 & 6.23 & 0.55 \\
        Full-Finetune & 1.80 & 1.52 & 4.79 & 0.030 & 7.65 & 0.75 \\
        LoRA-Finetune & 1.22 & 1.06 & 3.73 & 0.023 & 9.50 & 0.80 \\
        \textbf{\MethodName{} (ours)} & 1.34 & 1.07 & 3.84 & 0.029 & 6.44 & 3.30 \\
        \bottomrule
    \end{tabular}
    \caption{Per-user standard deviation of the real-world mocap calibration results.}
    \label{tab:real_mocap_user_std}
\end{table*}

\subsection{Mocap Device Selection}
Several glove-based hand motion capture systems are currently available, including professional EMF-tracking gloves such as MANUS Metagloves Pro, IMU/EMF-based products such as Rokoko Smartgloves, stretch-sensor gloves such as StretchSense gloves, and bend-sensor data gloves such as CyberGlove II~\cite{manus_metagloves_pro, rokoko_smartgloves, stretchsense_gloves, cyberglove_ii}. 
In our real-world experiment, we use OpenCyberGlove, a low-cost tensile-sensor-based glove~\citep{cyberorigin_opencyberglove}. 
This choice is motivated by both practicality and evaluation purpose: compared with high-end commercial mocap gloves, OpenCyberGlove provides an affordable platform for real data collection, while its tensile/strain sensor readings are naturally affected by user-specific hand geometry, glove fit, hand size, and hand shape. 
Therefore, it provides a representative and challenging test case for our calibration setting, where the goal is precisely to compensate for person- and fit-dependent distortions and recover accurate hand motion from limited paired calibration data.
\label{Appendix:Mocap Device}

\subsection{Additional K-Sweep Results}
\label{Appendix:K-Sweep}
\begin{table*}[h]
    \centering
    \setlength{\tabcolsep}{2pt}
    \begin{tabular}{llrrrrr}
        \toprule
        Method & $K$ & MPJPE$\downarrow$ & PA-MPJPE$\downarrow$ & Tip-PA$\downarrow$ & AUCJ$\uparrow$ & Pinch L2$\downarrow$ \\
          &  & (mm) & (mm) & (mm) &  & (mm) \\
        \midrule
        \MethodName{} (ours) & 1 & 16.07 & 10.94 & 24.40 & 0.694 & 38.04 \\
           & 2 & 16.10 & 10.86 & 24.30 & 0.694 & 36.66 \\
           & 4 & 15.64 & 10.45 & 23.29 & 0.702 & 34.90 \\
           & 8 & 15.21 & 10.19 & 22.26 & 0.709 & 33.87 \\
           & 16 & 15.10 & 10.14 & 21.94 & 0.712 & 33.70 \\
        \midrule
        LoRA FT & 1 & 18.96 & 14.07 & 31.33 & 0.651 & 43.44 \\
           & 2 & 17.80 & 13.00 & 28.54 & 0.671 & 42.01 \\
           & 4 & 16.38 & 11.67 & 25.33 & 0.694 & 37.28 \\
           & 8 & 15.44 & 10.65 & 22.92 & 0.709 & 33.42 \\
           & 16 & 15.12 & 10.29 & 22.06 & 0.714 & 32.89 \\
        \midrule
        Last-layer FT & 1 & 19.10 & 14.22 & 31.64 & 0.649 & 43.54 \\
           & 2 & 18.04 & 13.34 & 29.20 & 0.669 & 42.57 \\
           & 4 & 16.62 & 12.01 & 25.92 & 0.692 & 39.18 \\
           & 8 & 15.54 & 10.84 & 23.16 & 0.709 & 35.72 \\
           & 16 & 15.18 & 10.50 & 22.26 & 0.715 & 35.10 \\
        \midrule
        Ridge & 1 & 20.18 & 15.24 & 33.87 & 0.633 & 47.44 \\
           & 2 & 19.10 & 14.35 & 31.66 & 0.654 & 48.25 \\
           & 4 & 17.62 & 13.00 & 28.28 & 0.676 & 45.35 \\
           & 8 & 16.76 & 12.13 & 26.44 & 0.688 & 46.35 \\
           & 16 & 16.07 & 11.47 & 24.87 & 0.698 & 45.03 \\
        \midrule
        Full FT & 1 & 25.62 & 20.20 & 48.08 & 0.570 & 48.71 \\
           & 2 & 22.79 & 17.26 & 40.76 & 0.607 & 45.11 \\
           & 4 & 18.11 & 12.77 & 29.21 & 0.668 & 37.11 \\
           & 8 & 15.39 & 10.17 & 22.39 & 0.706 & 32.60 \\
           & 16 & 14.55 & 9.47 & 20.39 & 0.722 & 30.88 \\
        \bottomrule
    \end{tabular}
    \caption{Detailed K-sweep results on the augmented InterHand2.6M evaluation setting. All rows use the overall split aggregated over 3 seeds and 20 augmented units. }
    \label{tab:interhand_num_pairs_appendix}
\end{table*}

\Needspace{0.5\textheight}
\subsection{Additional Teleoperation Statistics}
\label{Appendix:Additional Teleoperation Statistics}
\begin{table}[htbp]
    \centering
    \footnotesize
    \setlength{\tabcolsep}{5pt}
    \resizebox{\linewidth}{!}{%
    \begin{tabular}{ccccc}
        \toprule
        Method & Bottle Cap Opening & Earbud Pinch Grasping & 3-Finger Glue Grasping & Tissue Box Opening \\
        & Avg. Time (s)$\downarrow$ & Avg. Time (s)$\downarrow$ & Avg. Time (s)$\downarrow$ & Avg. Time (s)$\downarrow$ \\
        \midrule
        No calibration & 37.9 & 7.0 & 10.5 & 42.0 \\
        LoRA-Finetune & 37.5 & 6.9 & \textbf{8.1} & 19.0 \\
        \textbf{\MethodName{} (ours)} & \textbf{27.3} & \textbf{4.8} & 9.9 & \textbf{17.4} \\
        \midrule
        Longest trial & 59.0 & 7.0 & 14.0 & 42.0 \\
        \bottomrule
    \end{tabular}%
    }
    \caption{Detailed downstream teleoperation completion-time statistics. Values are average completion time in seconds over 10 trials; lower is better. The longest-trial row reports the longest observed trial time for each task.}
    \label{tab:teleop_detail_statistics}
\end{table}

\subsection{Device and Compute Platform}
\label{Appendix:ComputePlatform}
All training runs and timing measurements are performed on a single-GPU server.
Table~\ref{tab:compute_platform} summarizes the hardware and operating system
used for model training and timing experiments.

\begin{table*}[ht]
    \centering
    \footnotesize
    \setlength{\tabcolsep}{6pt}
    \begin{tabular}{lp{0.40\linewidth}}
        \toprule
        Component & Value \\
        \midrule
        CPU & 2$\times$ Intel Xeon Platinum 8468V \\
        RAM & 2.0~TiB \\
        GPU & 1$\times$ NVIDIA H100 80~GB HBM3 \\
        \bottomrule
    \end{tabular}
    \caption{Device and compute platform used for training and timing experiments.}
    \label{tab:compute_platform}
\end{table*}

\subsection{Training Hyperparameters}
\label{Appendix:TrainingHyperparameters}
Table~\ref{tab:training_hyperparameters} reports the main optimization,
HyperNet, LoRA, and paired-pose preference-extractor hyperparameters used in
our training runs.

\begin{table*}[ht]
    \centering
    \setlength{\tabcolsep}{3pt}
    \begin{tabular}{p{0.55\linewidth}p{0.20\linewidth}}
        \toprule
        Setting & Value \\
        \midrule
        Seed & 42 \\
        Optimizer & Adam \\
        Total steps & 10{,}000 \\
        Batch size & 1024 \\
        Peak learning Rate & \texttt{3e-4} \\
        Final Learning Rate & \texttt{1e-6} \\
        Warmup Steps & 500 \\
        Gradient clipping & 1.0 \\
        LoRA rank / alpha & Rank 8; alpha 8.0 \\
        LoRA dropout & 0.1 \\
        Calibration Mapper Transformer Depth & 6 \\
        Calibration Mapper Transformer Number of Heads & 8 \\
        Decoder Flow Steps & 8 \\
        Decoder DIT Number of Layers & 4 \\
        Decoder DIT Number of Heads & 4 \\
        Hidden Dimension & 256 \\
        \bottomrule
    \end{tabular}
    \caption{Training hyperparameters used by \MethodName{}.}
    \label{tab:training_hyperparameters}
\end{table*}

\clearpage

\end{document}